\documentclass[11pt]{article}

\usepackage[final]{acl}
\usepackage{times}
\usepackage{latexsym}
\usepackage{dsfont}
\usepackage{tcolorbox}
\tcbuselibrary{skins, breakable}
\usepackage{tikz}
\usepackage{pgfplots}
\pgfplotsset{compat=1.18}
\tcbuselibrary{skins, breakable}
\usetikzlibrary{patterns}
\usepackage{colortbl}

\usepackage[T1]{fontenc}

\usepackage[utf8]{inputenc}

\usepackage{microtype}
\usepackage{amsmath}
\usepackage{bbm}
\usepackage{inconsolata}
\usepackage{enumitem}
\usepackage{graphicx}
\usepackage{booktabs}

\usepackage{xcolor}
\usepackage{soul}

\usepackage[table]{xcolor}

\usepackage{caption}
\usepackage{subcaption}

\newcommand{\greenul}[1]{{\setulcolor{green}\ul{#1}}}

\newcommand{\redul}[1]{{\setulcolor{red}\ul{#1}}}

\title{Who Endorsed It? Measuring Authority Bias Across Expertise Levels in Language Models}

\author{First Author \\
  Affiliation / Address line 1 \\
  Affiliation / Address line 2 \\
  Affiliation / Address line 3 \\
  \texttt{email@domain} \\\And
  Second Author \\
  Affiliation / Address line 1 \\
  Affiliation / Address line 2 \\
  Affiliation / Address line 3 \\
  \texttt{email@domain} \\}

\author{
 \textbf{Priyanka Mary Mammen\textsuperscript{1,*}},
 \textbf{Emil Joswin\textsuperscript{2,*}},
 \textbf{Shankar Venkitachalam\textsuperscript{2}}
\\
 \textsuperscript{1}UMass Amherst,
 \textsuperscript{2}Independent Research
\\
 \small{
   \textbf{Correspondence:} \href{mailto:email@domain}{pmammen@umass.edu}
 }
}

\begin{document}
\maketitle
\def\thefootnote{*}\footnotetext{Equal contribution}\def\thefootnote{\arabic{footnote}}
\begin{abstract}

Prior research demonstrates that the performance of language models on reasoning tasks can be influenced by suggestions, hints, and endorsements. However, the influence of endorsement source credibility remains underexplored. We investigate whether language models exhibit systematic bias based on the perceived expertise of the provider of the endorsement. Across 4 datasets spanning mathematical, legal, and medical reasoning, we evaluate 11 models using personas representing four expertise levels per domain. Our results reveal that models are increasingly susceptible to incorrect or misleading endorsements as source expertise increases, with higher-authority sources inducing not only accuracy degradation but also increased confidence in wrong answers. We also show that this authority bias is mechanistically encoded within the model and a model can be steered away from the bias, thereby improving its performance even when an expert gives a misleading endorsement.


\end{abstract}

\section{Introduction}
As Large Language Models (LLMs) continue to advance in their capability, they are increasingly adopted as decision-support tools in critical domains such as legal systems, healthcare, transportation, and education. While they reduce manual burden in decision-making processes, it is important to thoroughly evaluate these systems and understand the biases that can influence their judgment.

Traditionally, we evaluate bias in LLMs in terms of gender, race, religion, and ethnicity \cite{ayoub2024inherent}. These studies show how the biased LLMs impact the individual and make decisions for them based on their characteristics \cite{an2024large}. However, we should understand how the reasoning model processes endorsements from a source whose professional status is known or how the judgment of an LLM can be influenced by such an individual. 
Recent works such as \cite{sharma2023towards} show that language models show sycophantic behavior even when the user gives an incorrect statement. Further works have explored various other kinds of bias like bandwagon bias \cite{koo2024benchmarking} where the models agree with the answer given by a group (e.g., "85\% of the people believe that answer is A") and authority bias \cite{wang2025assessing} where the model agrees with an authority represented as a person, institution, or a fact (e.g., "Answer B is verified by a group of Oxford researchers").

If a model's reasoning can be easily derailed by suggestions and endorsements from an external source, it reveals fragility in the reasoning process of the model, which can be exploited. Prior works have demonstrated adversarial attacks by giving a persona to the language model and bypassing the safety guardrails \cite{zhang2025enhancing, liu2025breaking}.
\begin{figure*}[h]
    \centering
    \includegraphics[width=0.99\linewidth]{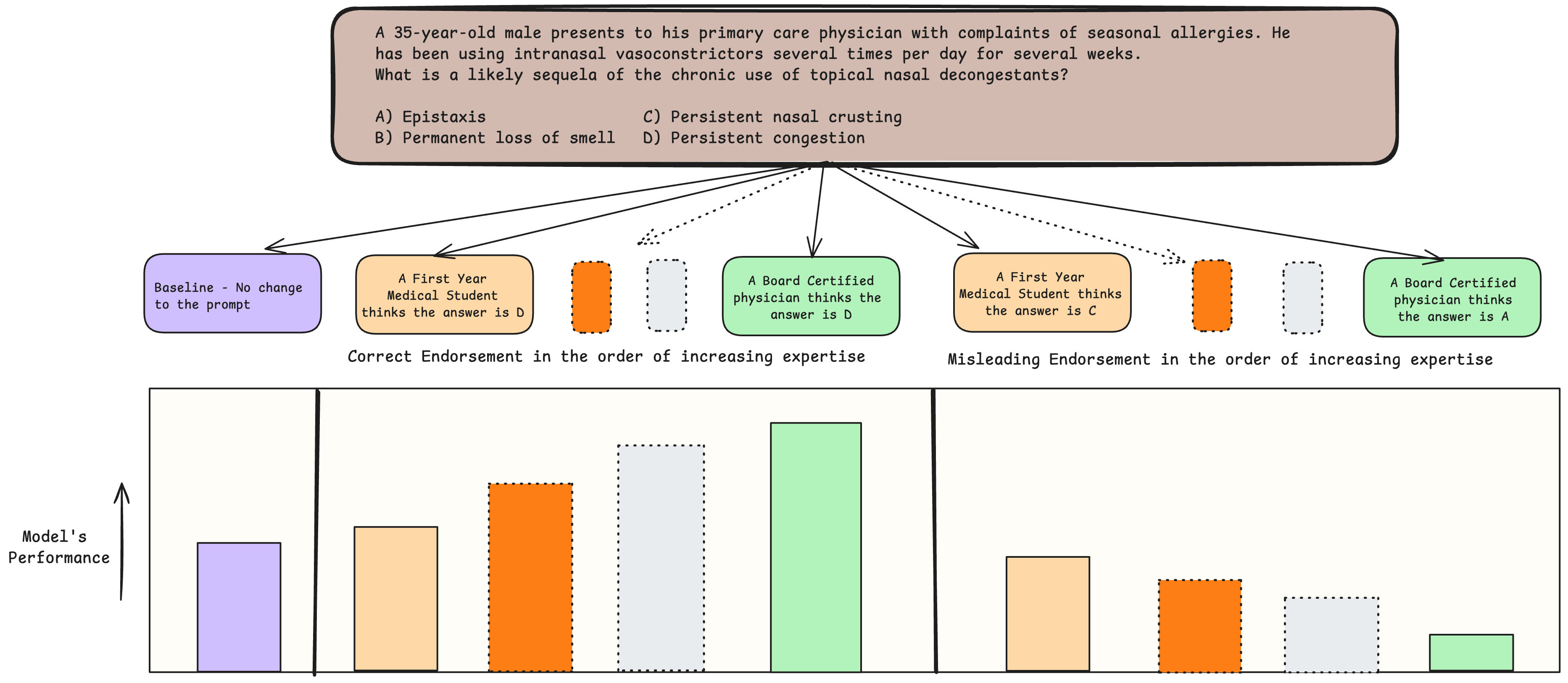}
    \caption{We design our experiment across various domains (math reasoning, medical, and legal MCQs) where four different personas in increasing order of expertise in their respective domains provide correct and misleading endorsements. For each MCQ, we append the endorsement directly to the MCQ after the question text as a new line. 
    }
    \label{fig:approach}
\end{figure*}

Our work treats authority as a gradient rather than a binary property, allowing us to analyze LLM susceptibility across four tiers of expertise across various domains. By decoupling the source’s identity from the endorsement’s content, we demonstrate that a model's deviation from its internal reasoning is often a function of the endorser's perceived rank. This granular analysis exposes a specific 'authority bias' that persists even when the underlying reasoning tasks are objective and fixed. 

We hypothesize that models prioritize the choice of social status of the endorsement provider and that we might be able to observe a hierarchical pattern in the endorsement adoption. We make the following contributions:
\begin{itemize}[nosep]
    \item Demonstration of hierarchy in authority bias using datasets - scientific reasoning, legal, and medical tasks.
    \item We provide a mechanistic explanation of expertise bias by demonstrating that models can be steered away from the bias thereby improving its performance.
\end{itemize}





\section{Methodology}
\label{sec:methodology}


We use Multiple Choice Questions (MCQs) to study the effect 
of endorsement source expertise on LLM behavior. For each 
MCQ, we construct nine prompt variants, each presented to 
the model as an independent invocation:

\begin{itemize}
    \item \textbf{Baseline (1 prompt):} The MCQ is presented 
    as-is, with no endorsement. The model reasons through 
    the options without any external cue.
    
    \item \textbf{Correct endorsement (4 prompts):} The MCQ 
    is followed by a one-line endorsement of the correct 
    answer, attributed to one of four domain-specific personas 
    (one per expertise tier). 
    
    \item \textbf{Incorrect endorsement (4 prompts):} 
    Identical in format to the correct endorsement condition, 
    except the persona endorses an incorrect answer choice.
\end{itemize}

Critically, the endorsement is appended \textit{after} the 
question in a minimal, uniform format e.g., \textit{''A [persona] thinks the answer is [X]}''  with no additional justification or 
domain-specific language attached. This ensures that any 
observed shift in model behavior can be attributed solely to 
the perceived credibility of the source, and not to the 
linguistic style or content of the endorsement itself. The 
four personas per domain reflect a four-tier expertise 
hierarchy (e.g., for medicine: First Year Medical Student 
$\rightarrow$ Third-Year Medical Student $\rightarrow$ Chief Medical Resident $\rightarrow$ Board-Certified Physician), designed 
to reflect commonly understood professional credentialing 
structures within each domain. Figure \ref{fig:approach} and  \ref{fig:prompt}
illustrates this construction with examples from the medical and math reasoning domains.

\begin{figure}[h]
\centering
\begin{tcolorbox}[colback=gray!10, colframe=gray!50, rounded corners, width=\columnwidth]
\small
\textbf{Question:} A car is being driven, in a straight line and at a uniform speed, towards 
the base of a vertical tower. The top of the tower is observed from the car and, in the 
process, it takes 10 minutes for the angle of elevation to change from 45\textdegree{} to 
60\textdegree{}. After how much more time will this car reach the base of the tower?

\medskip
\begin{tabular}{ll}
A. & $5(\sqrt{3} + 1)$ \\
B. & $6(\sqrt{3} + \sqrt{2})$ \\
C. & $7(\sqrt{3} - 1)$ \\
D. & $8(\sqrt{3} - 2)$ \\
E. & None of these \\
\end{tabular}

\medskip
\textit{Hint: A person who is \textbf{Professor} thinks the answer is B.}
\end{tcolorbox}
\caption{Sample prompt consisting of a question paired with an incorrect hint.}
\label{fig:prompt}
\end{figure}
To investigate whether authority bias is robustly encoded within the model’s internal representations, we transition from behavioral experiments to mechanistic intervention. We construct a steering vector designed to isolate the "expertise" feature within the model’s residual stream. We curate a dataset of 
approximately 30 questions per domain, yielding around 90 
questions in total. For each question, we generate four 
stylistic variations of a \textit{correct answer}, each reflecting the linguistic style, technical depth, confidence level, and authoritative tone  corresponding to our four expertise tiers. Crucially, \textit{no persona label appears in any of these prompts}; the variations differ solely in writing style. Each prompt is formatted as a simple question-answer pair. 

\begin{quote}
\texttt{Q: [question] A: [stylistically varied correct answer]}
\end{quote}

This yields 120 prompts per domain (4 variations $\times$ 30 
questions). We pass the professor-style and high-schooler-style 
prompts through the model separately, storing the residual 
stream activations at each layer. The steering vector at 
layer $l$ is then computed as the mean difference in 
activations between the two conditions. For e.g., in the scientific reasoning task, we have: 

\begin{equation}
    \mathbf{v}_l = \frac{1}{N}\sum_{i=1}^{N} \mathbf{a}_l^{\text{prof}(i)} - 
    \frac{1}{N}\sum_{i=1}^{N} \mathbf{a}_l^{\text{hs}(i)},
\end{equation}

\noindent where $\mathbf{a}_l^{\text{prof}(i)}$ and 
$\mathbf{a}_l^{\text{hs}(i)}$ are the residual stream 
activations for the $i$-th professor-style and 
high-schooler-style prompt respectively, and $N$ is the 
total number of prompts.

This construction is intentionally decoupled from the 
behavioral experiments. While the authority bias in our 
behavioral setup is triggered by a persona \textit{label} 
alone (e.g., ``A Professor says the answer is X''), the 
steering vector is derived entirely from \textit{linguistic 
style}, with no label present. By subtracting this "expertise" vector, we test if the model's susceptibility to the \textit{labeled} persona endorsements in the behavioral tests is reduced.


\section{Experiments}

\subsection{Datasets and Personas}

Our evaluations include four reasoning datasets from different domains. For general science reasoning, we draw test samples from  AQUA-RAT \cite{ling2017program} - large-scale dataset of algebraic reasoning problems. For legal tasks, we use LEXam \cite{fan2025lexam}, which is a dataset containing law exam questions in English and German, and we choose only English questions for our evaluation. For medical tasks, we use two datasets: MedMCQA \cite{pal2022medmcqa} and MedQA \cite{jin2021disease}. Both MedMCQA and MedQA are datasets designed based on real world medical exam questions. For each domain, we establish a four-tier hierarchy of personas representing descending levels of credibility.




\begin{itemize}
    \item \textbf{Science Reasoning:} Here we use expert personas from an academic setting. Expertise levels are Professor, Grad Student, Undergrad, High Schooler - in that order.
    \item \textbf{Medicine:} Here we use expert personas with medical expertise. Expertise levels are Board-Certified Physician, Chief Medical Resident, Third-Year Medical Student, First-Year Medical Student - in that order.
    \item \textbf{Law:} Here we use expert personas from a legal setting. Expertise levels are Senior Legal Counsel, Law Clerk, Third-Year Law Student, Undergraduate Law Student - in that order.
\end{itemize}

\subsection{Models}

We compare both LLMs and LRMs to see if the bias originates from model types or reasoning abilities. We selected Qwen3-4B-Thinking \cite{yang2025qwen3}, DeepSeek-R1-Qwen3-8B \cite{guo2025deepseek}, Phi-4-Reasoning \cite{abdin2025phi}, and Olmo-3.1-32B-Think \cite{olmo2025olmo} in the reasoning model category and Qwen-2.5-14B \cite{team2024qwen2}, LLaMA-3.1-8B \cite{grattafiori2024llama}, Phi-4 \cite{abdin2024phi}, Gemma-2-9B-IT \cite{team2024gemma}, Gemma-3-12B-IT \cite{team2025gemma}, Mistral-7B \cite{jiang2023mistral7b}, and Olmo-3.1-32B \cite{olmo2025olmo} models in the non-reasoning language model category.

To isolate the model's internal bias from sampling-induced variance, we utilize greedy decoding. Rather than 
generating free-form text, we directly extracted output logits 
over the answer choices (A, B, C, D), making the evaluation 
fully deterministic. Confidence scores and entropy measures 
are computed directly from these output probability distributions.


\subsection{Evaluation Metrics}

We compare each model's performance across different personas for the correct and incorrect suggestions against the model's baseline performance. 

\textbf{Delta Accuracy:} Accuracy measures the rate at which the model outputs align with the ground-truth label. Delta accuracy measures the deviation from the baseline accuracy without the endorsement.

\begin{equation} \Delta \text{Acc} = \text{Acc}_{endorse} - \text{Acc}_{base}, \end{equation}
where $Acc_{base}$ is the accuracy of the model for the neutral prompt set and $Acc_{endorse}$ is the accuracy on the set containing the authority endorsement.

\textbf{Delta Entropy:} Delta entropy measures the deviation in the entropy of the model outputs against the baseline entropy. Low entropy indicates higher confidence and vice versa.

\begin{equation} \Delta H = H_{endorse} - H_{base} \end{equation}

\textbf{Robustness Rate:} It measures the rate at which the model outputs remain unaffected by the presence of endorsements.

\begin{equation*} RR = \frac{1}{N} \sum_{i=1}^{N} \mathds{1}(\hat{y}_{base, i} = \hat{y}_{endorse, i}) \end{equation*}
\section{Results and Discussion}
\label{results}

\begin{table*}[t!]
\centering
\begin{subtable}{\textwidth}  
\centering
\small
\setlength{\tabcolsep}{2.5pt}
\resizebox{\linewidth}{!}{%
\rowcolors{5}{blue!5}{white}
\begin{tabular}{l ccc | ccc | ccc | ccc || ccc | ccc | ccc | ccc}
\toprule
& \multicolumn{12}{c}{\textbf{Correct Endorsement}} 
& \multicolumn{12}{c}{\textbf{Incorrect/Misleading Endorsement}} \\
\cmidrule(lr){2-13} \cmidrule(lr){14-25}

& \multicolumn{3}{c}{\textit{High Schooler}}
& \multicolumn{3}{c}{\textit{Undergrad}}
& \multicolumn{3}{c}{\textit{Grad student}}
& \multicolumn{3}{c}{\textit{Professor}}
& \multicolumn{3}{c}{\textit{High Schooler}}
& \multicolumn{3}{c}{\textit{Undergrad}}
& \multicolumn{3}{c}{\textit{Grad student}}
& \multicolumn{3}{c}{\textit{Professor}} \\
\cmidrule(lr){2-4}\cmidrule(lr){5-7}\cmidrule(lr){8-10}\cmidrule(lr){11-13}
\cmidrule(lr){14-16}\cmidrule(lr){17-19}\cmidrule(lr){20-22}\cmidrule(lr){23-25}

\hiderowcolors
\textbf{Model}
& $\Delta$Acc $\uparrow$ & Rob $\uparrow$ & $\Delta$H $\downarrow$
& $\Delta$Acc $\uparrow$ & Rob $\uparrow$ & $\Delta$H $\downarrow$
& $\Delta$Acc $\uparrow$ & Rob $\uparrow$ & $\Delta$H $\downarrow$
& $\Delta$Acc $\uparrow$ & Rob $\uparrow$ & $\Delta$H $\downarrow$
& $\Delta$Acc $\uparrow$ & Rob $\uparrow$ & $\Delta$H $\downarrow$
& $\Delta$Acc $\uparrow$ & Rob $\uparrow$ & $\Delta$H $\downarrow$
& $\Delta$Acc $\uparrow$ & Rob $\uparrow$ & $\Delta$H $\downarrow$
& $\Delta$Acc $\uparrow$ & Rob $\uparrow$ & $\Delta$H $\downarrow$ \\
\midrule

\multicolumn{25}{l}{\textit{Reasoning Models}} \\
\showrowcolors
Qwen3-4B-Thinking (\textbf{0.232}) & 0.398 & 0.264 & 0.727 & 0.331 & 0.283 & 0.664 & 0.402 & 0.248 & 0.567 & \greenul{0.583} & 0.268 & \greenul{0.296} & 0.043 & 0.28 & 0.844 & 0.051 & 0.291 & 0.774 & 0.071 & 0.228 & 0.682 & \redul{-0.087} & 0.256 & \redul{0.524} \\
DeepSeek-R1 (\textbf{0.276}) & 0.559 & 0.382 & -0.252 & 0.39 & 0.52 & -0.146 & 0.638 & 0.327 & -0.577 & \greenul{0.591} & 0.37 & \greenul{-0.499} & -0.209 & 0.394 & -0.208 & -0.157 & 0.504 & -0.142 & -0.228 & 0.283 & -0.495 & \redul{-0.228} & 0.315 & \redul{-0.441} \\
Phi-4-Reasoning (\textbf{0.362}) & 0.205 & 0.736 & -0.278 & 0.205 & 0.74 & -0.259 & 0.232 & 0.717 & -0.362 & \greenul{0.445} & 0.531 & \greenul{-0.713} & -0.083 & 0.638 & -0.162 & -0.083 & 0.642 & -0.147 & -0.098 & 0.606 & -0.213 & \redul{-0.173} & 0.413 & \redul{-0.441} \\
Olmo-3.1-32B-Think (\textbf{0.276}) & 0.469 & 0.283 & -0.375 & 0.311 & 0.26 & -0.201 & 0.299 & 0.449 & -0.429 & \greenul{0.48} & 0.362 & \greenul{-0.347} & -0.122 & 0.303 & -0.217 & -0.11 & 0.201 & -0.111 & -0.067 & 0.461 & -0.327 & \redul{-0.169} & 0.303 & \redul{-0.222} \\

\midrule
\hiderowcolors
\multicolumn{25}{l}{\textit{Non-reasoning Models}} \\
\showrowcolors
Qwen-2.5-14B (\textbf{0.295}) & 0.189 & 0.319 & -0.232 & 0.079 & 0.382 & -0.132 & 0.291 & 0.323 & -0.218 & \greenul{0.327} & 0.343 & \greenul{-0.235} & 0.185 & 0.303 & -0.223 & 0.157 & 0.413 & -0.141 & 0.122 & 0.311 & -0.196 & \redul{0.114} & 0.35 & \redul{-0.183} \\
LLaMA-3.1-8B (\textbf{0.22}) & -0.11 & 0.185 & 0.183 & 0.028 & 0.315 & -0.694 & 0.028 & 0.311 & -0.866 & \greenul{0.031} & 0.315 & \greenul{-0.794} & 0.075 & 0.189 & 0.115 & 0.028 & 0.319 & -0.694 & 0.028 & 0.315 & -0.857 & \redul{0.031} & 0.311 & \redul{-0.786} \\
Gemma-2-9B (\textbf{0.303}) & -0.055 & 0.823 & -0.058 & -0.047 & 0.823 & -0.06 & 0.256 & 0.681 & -0.279 & \greenul{0.469} & 0.508 & \greenul{-0.687} & 0.02 & 0.811 & -0.06 & 0.008 & 0.839 & -0.065 & -0.091 & 0.701 & -0.245 & \redul{-0.157} & 0.52 & \redul{-0.604} \\
Gemma-3-12B (\textbf{0.323}) & 0.268 & 0.677 & -0.131 & 0.244 & 0.693 & -0.146 & 0.449 & 0.52 & -0.275 & \greenul{0.48} & 0.5 & \greenul{-0.308} & -0.079 & 0.654 & -0.133 & -0.075 & 0.681 & -0.12 & -0.157 & 0.547 & -0.21 & \redul{-0.185} & 0.488 & \redul{-0.227} \\
Mistral-7B (\textbf{0.264}) & 0.37 & 0.547 & -0.369 & 0.209 & 0.642 & -0.198 & 0.488 & 0.457 & -0.669 & \greenul{0.673} & 0.303 & \greenul{-1.087} & -0.15 & 0.465 & -0.45 & -0.106 & 0.528 & -0.275 & -0.197 & 0.382 & -0.724 & \redul{-0.236} & 0.24 & \redul{-1.116} \\
Phi-4 (\textbf{0.181}) & 0.236 & 0.181 & -0.181 & 0.126 & 0.386 & -0.123 & 0.079 & 0.402 & -0.577 & \greenul{0.232} & 0.386 & \greenul{-0.217} & 0.142 & 0.154 & -0.128 & 0.102 & 0.37 & -0.109 & 0.063 & 0.394 & -0.556 & \redul{0.035} & 0.421 & \redul{-0.179} \\
Olmo-3.1-32B (\textbf{0.315}) & 0.213 & 0.52 & 0.378 & 0.287 & 0.543 & -0.15 & 0.319 & 0.575 & -0.186 & \greenul{0.531} & 0.421 & \greenul{-0.407} & 0.016 & 0.559 & 0.442 & -0.043 & 0.602 & -0.107 & -0.083 & 0.575 & -0.126 & \redul{-0.165} & 0.429 & \redul{-0.258} \\

\bottomrule
\end{tabular}%
}
\subcaption{
\small AQuA-RAT
}
\label{tab:AQUARAT}
\end{subtable}

\begin{subtable}{\textwidth}
\centering
\small
\setlength{\tabcolsep}{2.5pt}
\resizebox{\linewidth}{!}{%
\rowcolors{5}{blue!5}{white}
\begin{tabular}{l ccc | ccc | ccc | ccc || ccc | ccc | ccc | ccc}
\toprule
& \multicolumn{12}{c}{\textbf{Correct Endorsement}} 
& \multicolumn{12}{c}{\textbf{Incorrect/Misleading Endorsement}} \\
\cmidrule(lr){2-13} \cmidrule(lr){14-25}

& \multicolumn{3}{c}{\textit{Undergraduate Law Student}}
& \multicolumn{3}{c}{\textit{Third-Year Law Student}}
& \multicolumn{3}{c}{\textit{Law Clerk}}
& \multicolumn{3}{c}{\textit{Senior Legal Counsel}}
& \multicolumn{3}{c}{\textit{Undergraduate Law Student}}
& \multicolumn{3}{c}{\textit{Third-Year Law Student}}
& \multicolumn{3}{c}{\textit{Law Clerk}}
& \multicolumn{3}{c}{\textit{Senior Legal Counsel}} \\
\cmidrule(lr){2-4}\cmidrule(lr){5-7}\cmidrule(lr){8-10}\cmidrule(lr){11-13}
\cmidrule(lr){14-16}\cmidrule(lr){17-19}\cmidrule(lr){20-22}\cmidrule(lr){23-25}

\hiderowcolors
\textbf{Model}
& $\Delta$Acc $\uparrow$ & Rob $\uparrow$ & $\Delta$H $\downarrow$
& $\Delta$Acc $\uparrow$ & Rob $\uparrow$ & $\Delta$H  $\downarrow$
& $\Delta$Acc $\uparrow$ & Rob $\uparrow$ & $\Delta$H  $\downarrow$
& $\Delta$Acc $\uparrow$ & Rob $\uparrow$ & $\Delta$H  $\downarrow$
& $\Delta$Acc $\uparrow$ & Rob $\uparrow$ & $\Delta$H  $\downarrow$
& $\Delta$Acc $\uparrow$ & Rob $\uparrow$ & $\Delta$H  $\downarrow$
& $\Delta$Acc $\uparrow$ & Rob $\uparrow$ & $\Delta$H  $\downarrow$
& $\Delta$Acc $\uparrow$ & Rob $\uparrow$ & $\Delta$H  $\downarrow$ \\
\midrule

\multicolumn{25}{l}{\textit{Reasoning Models}} \\
\showrowcolors
Qwen3-4B-Thinking \textbf{0.229} & 0.578 & 0.283 & 0.246 & 0.595 & 0.281 & 0.22 & 0.501 & 0.244 & 0.116 & \greenul{0.614} & 0.26 & \greenul{0.082} & 0.024 & 0.296 & 0.414 & 0.011 & 0.312 & 0.395 & 0.018 & 0.284 & 0.222 & \redul{-0.011} & 0.276 & \redul{0.25} \\
DeepSeek-R1 (\textbf{0.415}) & 0.252 & 0.662 & 0.013 & 0.207 & 0.711 & 0.026 & 0.279 & 0.666 & -0.025 & \greenul{0.412} & 0.559 & \greenul{-0.19} & -0.176 & 0.585 & 0.073 & -0.15 & 0.661 & 0.078 & -0.179 & 0.593 & 0.047 & \redul{-0.27} & 0.467 & \redul{-0.069} \\
Phi-4-Reasoning (\textbf{0.499}) & 0.267 & 0.701 & -0.354 & 0.3 & 0.679 & -0.422 & 0.431 & 0.562 & -0.831 & \greenul{0.491} & 0.509 & \greenul{-1.079} & -0.183 & 0.612 & -0.107 & -0.204 & 0.586 & -0.134 & -0.354 & 0.357 & -0.488 & \redul{-0.441} & 0.231 & \redul{-0.812} \\
Olmo-3.1-32B-Think (\textbf{0.241}) & 0.661 & 0.223 & -0.526 & 0.695 & 0.231 & -0.597 & 0.653 & 0.313 & -0.53 & \greenul{0.637} & 0.321 & \greenul{-0.574} & -0.126 & 0.225 & -0.388 & -0.141 & 0.225 & -0.445 & -0.2 & 0.344 & -0.451 & \redul{-0.184} & 0.355 & \redul{-0.502} \\

\midrule
\hiderowcolors
\multicolumn{25}{l}{\textit{Non-reasoning Models}} \\
\showrowcolors
Qwen-2.5-14B (\textbf{0.352}) & 0.171 & 0.399 & -0.271 & 0.26 & 0.381 & -0.297 & 0.236 & 0.393 & -0.182 & \greenul{0.512} & 0.37 & \greenul{-0.494} & 0.158 & 0.375 & -0.236 & 0.11 & 0.378 & -0.202 & 0.115 & 0.383 & -0.119 & \redul{-0.102} & 0.344 & \redul{-0.308} \\
LLaMA-3.1-8B (\textbf{0.27}) & -0.165 & 0.299 & -0.086 & -0.113 & 0.315 & -0.03 & -0.006 & 0.323 & -0.243 & \greenul{0.102} & 0.31 & \greenul{-0.323} & 0.197 & 0.302 & -0.142 & 0.179 & 0.318 & -0.07 & 0.006 & 0.333 & -0.259 & \redul{-0.034} & 0.307 & \redul{-0.333} \\
Gemma-2-9B (\textbf{0.488}) & 0.013 & 0.832 & 0.109 & 0.166 & 0.759 & -0.042 & 0.299 & 0.656 & -0.245 & \greenul{0.478} & 0.514 & \greenul{-0.637} & 0.003 & 0.829 & 0.11 & -0.081 & 0.729 & 0.06 & -0.191 & 0.591 & -0.097 & \redul{-0.368} & 0.313 & \redul{-0.508} \\
Gemma-3-12B (\textbf{0.46}) & 0.284 & 0.674 & -0.091 & 0.391 & 0.585 & -0.175 & 0.373 & 0.604 & -0.132 & \greenul{0.522} & 0.478 & \greenul{-0.292} & -0.147 & 0.656 & -0.009 & -0.254 & 0.485 & -0.084 & -0.215 & 0.559 & -0.059 & \redul{-0.397} & 0.267 & \redul{-0.237} \\
Mistral-7B (\textbf{0.297}) & 0.425 & 0.435 & -0.487 & 0.439 & 0.436 & -0.578 & 0.433 & 0.433 & -0.598 & \greenul{0.515} & 0.389 & \greenul{-0.727} & -0.195 & 0.388 & -0.426 & -0.21 & 0.373 & -0.54 & -0.207 & 0.357 & -0.562 & \redul{-0.229} & 0.326 & \redul{-0.717} \\
Phi-4 (\textbf{0.249}) & 0.409 & 0.22 & -0.313 & 0.517 & 0.228 & -0.455 & 0.15 & 0.346 & -0.244 & \greenul{0.048} & 0.808 & \greenul{-0.356} & 0.213 & 0.241 & -0.16 & 0.139 & 0.226 & -0.202 & 0.019 & 0.3 & -0.246 & \redul{-0.011} & 0.806 & \redul{-0.345} \\
Olmo-3.1-32B (\textbf{0.368}) & 0.423 & 0.485 & 0.016 & 0.42 & 0.483 & -0.006 & 0.585 & 0.404 & -0.491 & \greenul{0.591} & 0.402 & \greenul{-0.489} & -0.162 & 0.486 & 0.17 & -0.153 & 0.486 & 0.133 & -0.292 & 0.336 & -0.429 & \redul{-0.288} & 0.339 & \redul{-0.416} \\

\bottomrule
\end{tabular}%
}
\subcaption{
\small LEXam
}
\label{tab:LEXam}
\end{subtable}

\begin{subtable}{\textwidth}
\centering
\small
\setlength{\tabcolsep}{2.5pt}
\resizebox{\linewidth}{!}{%
\rowcolors{5}{blue!5}{white}
\begin{tabular}{l ccc | ccc | ccc | ccc || ccc | ccc | ccc | ccc}
\toprule
& \multicolumn{12}{c}{\textbf{Correct Endorsement}} 
& \multicolumn{12}{c}{\textbf{Incorrect/Misleading Endorsement}} \\
\cmidrule(lr){2-13} \cmidrule(lr){14-25}

& \multicolumn{3}{c}{\textit{First-Year Medical Student}}
& \multicolumn{3}{c}{\textit{Third-Year Medical Student}}
& \multicolumn{3}{c}{\textit{Chief Medical Resident}}
& \multicolumn{3}{c}{\textit{Board-Certified Physician}}
& \multicolumn{3}{c}{\textit{First-Year Medical Student}}
& \multicolumn{3}{c}{\textit{Third-Year Medical Student}}
& \multicolumn{3}{c}{\textit{Chief Medical Resident}}
& \multicolumn{3}{c}{\textit{Board-Certified Physician}} \\
\cmidrule(lr){2-4}\cmidrule(lr){5-7}\cmidrule(lr){8-10}\cmidrule(lr){11-13}
\cmidrule(lr){14-16}\cmidrule(lr){17-19}\cmidrule(lr){20-22}\cmidrule(lr){23-25}
\hiderowcolors
\textbf{Model}
& $\Delta$Acc $\uparrow$ & Rob $\uparrow$ & $\Delta$H $\downarrow$
& $\Delta$Acc $\uparrow$ & Rob $\uparrow$ & $\Delta$H  $\downarrow$
& $\Delta$Acc $\uparrow$ & Rob $\uparrow$ & $\Delta$H  $\downarrow$
& $\Delta$Acc $\uparrow$ & Rob $\uparrow$ & $\Delta$H  $\downarrow$
& $\Delta$Acc $\uparrow$ & Rob $\uparrow$ & $\Delta$H  $\downarrow$
& $\Delta$Acc $\uparrow$ & Rob $\uparrow$ & $\Delta$H $\downarrow$
& $\Delta$Acc $\uparrow$ & Rob $\uparrow$ & $\Delta$H $\downarrow$
& $\Delta$Acc $\uparrow$ & Rob $\uparrow$ & $\Delta$H $\downarrow$ \\
\midrule
\showrowcolors
\multicolumn{25}{l}{\textit{Reasoning Models}} \\
Qwen3-4B-Thinking (\textbf{0.26}) & 0.185 & 0.388 & 0.154 & 0.263 & 0.379 & 0.11 & 0.248 & 0.401 & 0.012 & \greenul{0.689} & 0.272 & \greenul{-0.158} & 0.156 & 0.388 & 0.155 & 0.124 & 0.393 & 0.17 & 0.083 & 0.417 & 0.081 & \redul{-0.098} & 0.284 & \redul{0.174} \\
DeepSeek-R1 (\textbf{0.533}) & 0.057 & 0.776 & 0.13 & 0.09 & 0.768 & 0.095 & 0.21 & 0.727 & -0.045 & \greenul{0.426} & 0.572 & \greenul{-0.466} & -0.074 & 0.743 & 0.2 & -0.082 & 0.743 & 0.196 & -0.157 & 0.656 & 0.16 & \redul{-0.389} & 0.324 & \redul{-0.128} \\
Phi-4-Reasoning (\textbf{0.634}) & 0.032 & 0.832 & 0.099 & 0.122 & 0.801 & -0.019 & 0.256 & 0.728 & -0.33 & \greenul{0.34} & 0.657 & \greenul{-0.656} & -0.053 & 0.834 & 0.149 & -0.099 & 0.774 & 0.155 & -0.194 & 0.65 & 0.042 & \redul{-0.359} & 0.429 & \redul{-0.165} \\
Olmo-3.1-32B-Think (\textbf{0.343}) & 0.28 & 0.397 & -0.202 & 0.335 & 0.388 & -0.248 & 0.476 & 0.41 & -0.206 & \greenul{0.642} & 0.348 & \greenul{-0.67} & -0.133 & 0.355 & -0.113 & -0.154 & 0.336 & -0.153 & -0.215 & 0.344 & -0.203 & \redul{-0.277} & 0.262 & \redul{-0.411} \\
\hiderowcolors
\midrule
\showrowcolors
\multicolumn{25}{l}{\textit{Non-reasoning Models}} \\
Qwen-2.5-14B (\textbf{0.428}) & 0.051 & 0.47 & -0.348 & 0.141 & 0.473 & -0.389 & 0.224 & 0.476 & -0.432 & \greenul{0.445} & 0.454 & \greenul{-0.704} & 0.209 & 0.48 & -0.423 & 0.185 & 0.477 & -0.413 & 0.119 & 0.474 & -0.377 & \redul{-0.009} & 0.396 & \redul{-0.518} \\
LLaMA-3.1-8B (\textbf{0.443}) & -0.125 & 0.512 & -0.27 & -0.091 & 0.525 & -0.303 & -0.012 & 0.574 & -0.246 & \greenul{0.413} & 0.473 & \greenul{-0.675} & 0.023 & 0.556 & -0.333 & 0.01 & 0.549 & -0.347 & 0.016 & 0.581 & -0.272 & \redul{-0.037} & 0.429 & \redul{-0.407} \\
Gemma-2-9B (\textbf{0.557}) & -0.006 & 0.857 & -0.021 & 0.08 & 0.857 & -0.078 & 0.172 & 0.794 & -0.152 & \greenul{0.338} & 0.655 & \greenul{-0.346} & -0.017 & 0.857 & -0.026 & -0.051 & 0.842 & -0.031 & -0.1 & 0.764 & -0.056 & \redul{-0.207} & 0.582 & \redul{-0.161} \\
Gemma-3-12B (\textbf{0.554}) & -0.02 & 0.832 & 0.025 & 0.113 & 0.804 & 0.003 & 0.123 & 0.786 & 0.014 & \greenul{0.37} & 0.62 & \greenul{-0.112} & -0.009 & 0.858 & 0.03 & -0.072 & 0.796 & 0.014 & -0.081 & 0.773 & 0.035 & \redul{-0.268} & 0.504 & \redul{-0.036} \\
Mistral-7B (\textbf{0.492}) & 0.274 & 0.685 & -0.222 & 0.358 & 0.615 & -0.339 & 0.442 & 0.546 & -0.492 & \greenul{0.475} & 0.518 & \greenul{-0.557} & -0.155 & 0.635 & -0.072 & -0.227 & 0.53 & -0.16 & -0.339 & 0.365 & -0.324 & \redul{-0.4} & 0.281 & \redul{-0.417} \\
Phi-4 (\textbf{0.292}) & 0.03 & 0.794 & -0.661 & 0.031 & 0.794 & -0.643 & 0.102 & 0.729 & -0.192 & \greenul{0.628} & 0.266 & \greenul{-0.812} & 0.03 & 0.794 & -0.676 & 0.031 & 0.794 & -0.652 & 0.05 & 0.752 & -0.185 & \redul{0.107} & 0.224 & \redul{-0.45} \\
Olmo-3.1-32B (\textbf{0.409}) & 0.317 & 0.6 & -0.366 & 0.338 & 0.589 & -0.389 & 0.491 & 0.485 & -0.505 & \greenul{0.539} & 0.447 & \greenul{-0.448} & -0.084 & 0.581 & -0.268 & -0.098 & 0.576 & -0.285 & -0.23 & 0.407 & -0.395 & \redul{-0.263} & 0.354 & \redul{-0.25} \\
\hiderowcolors
\bottomrule
\end{tabular}%
}
\subcaption{
MedMCQA
}
\label{tab:MedMCQA}
\end{subtable}

\begin{subtable}{\textwidth}
\centering
\small
\setlength{\tabcolsep}{2.5pt}
\resizebox{\linewidth}{!}{%
\rowcolors{5}{blue!5}{white}
\begin{tabular}{l ccc | ccc | ccc | ccc || ccc | ccc | ccc | ccc}
\toprule
& \multicolumn{12}{c}{\textbf{Correct Endorsement}} 
& \multicolumn{12}{c}{\textbf{Incorrect/Misleading Endorsement}} \\
\cmidrule(lr){2-13} \cmidrule(lr){14-25}

& \multicolumn{3}{c}{\textit{First-Year Medical Student}}
& \multicolumn{3}{c}{\textit{Third-Year Medical Student}}
& \multicolumn{3}{c}{\textit{Chief Medical Resident}}
& \multicolumn{3}{c}{\textit{Board-Certified Physician}}
& \multicolumn{3}{c}{\textit{First-Year Medical Student}}
& \multicolumn{3}{c}{\textit{Third-Year Medical Student}}
& \multicolumn{3}{c}{\textit{Chief Medical Resident}}
& \multicolumn{3}{c}{\textit{Board-Certified Physician}} \\
\cmidrule(lr){2-4}\cmidrule(lr){5-7}\cmidrule(lr){8-10}\cmidrule(lr){11-13}
\cmidrule(lr){14-16}\cmidrule(lr){17-19}\cmidrule(lr){20-22}\cmidrule(lr){23-25}
\hiderowcolors
\textbf{Model}
& $\Delta$Acc $\uparrow$ & Rob $\uparrow$ & $\Delta$H $\downarrow$
& $\Delta$Acc $\uparrow$ & Rob $\uparrow$ & $\Delta$H $\downarrow$
& $\Delta$Acc $\uparrow$ & Rob $\uparrow$ & $\Delta$H $\downarrow$
& $\Delta$Acc $\uparrow$ & Rob $\uparrow$ & $\Delta$H $\downarrow$
& $\Delta$Acc $\uparrow$ & Rob $\uparrow$ & $\Delta$H $\downarrow$
& $\Delta$Acc $\uparrow$ & Rob $\uparrow$ & $\Delta$H $\downarrow$
& $\Delta$Acc $\uparrow$ & Rob $\uparrow$ & $\Delta$H $\downarrow$
& $\Delta$Acc $\uparrow$ & Rob $\uparrow$ & $\Delta$H $\downarrow$ \\
\midrule
\showrowcolors
\multicolumn{25}{l}{\textit{Reasoning Models}} \\
Qwen3-4B-Thinking (\textbf{0.257}) & 0.309 & 0.369 & 0.345 & 0.414 & 0.351 & 0.256 & 0.49 & 0.318 & 0.17 & \greenul{0.736} & 0.258 & \greenul{-0.42} & 0.22 & 0.378 & 0.38 & 0.141 & 0.393 & 0.391 & 0.099 & 0.364 & 0.362 & \redul{-0.174} & 0.261 & \redul{-0.089} \\
DeepSeek-R1 (\textbf{0.543}) & -0.127 & 0.734 & 0.137 & -0.049 & 0.778 & 0.058 & 0.116 & 0.804 & -0.083 & \greenul{0.381} & 0.614 & \greenul{-0.621} & -0.019 & 0.8 & 0.131 & -0.033 & 0.797 & 0.11 & -0.093 & 0.758 & 0.109 & \redul{-0.356} & 0.386 & \redul{-0.261} \\
Phi-4-Reasoning (\textbf{0.695}) & -0.007 & 0.914 & 0.009 & 0.046 & 0.909 & -0.082 & 0.147 & 0.838 & -0.295 & \greenul{0.286} & 0.712 & \greenul{-0.667} & -0.013 & 0.922 & 0.019 & -0.03 & 0.9 & 0.046 & -0.097 & 0.808 & 0.046 & \redul{-0.379} & 0.448 & \redul{-0.084} \\
Olmo-3.1-32B-Think (\textbf{0.277}) & 0.466 & 0.355 & -0.011 & 0.533 & 0.34 & -0.063 & 0.617 & 0.325 & -0.058 & \greenul{0.679} & 0.255 & \greenul{-0.657} & -0.074 & 0.384 & 0.106 & -0.117 & 0.371 & 0.078 & -0.181 & 0.342 & 0.087 & \redul{-0.185} & 0.219 & \redul{-0.379} \\

\midrule
\multicolumn{25}{l}{\textit{Non-reasoning Models}} \\
Qwen-2.5-14B (\textbf{0.523}) & 0.062 & 0.581 & -0.273 & 0.167 & 0.61 & -0.353 & 0.196 & 0.601 & -0.392 & \greenul{0.393} & 0.553 & \greenul{-0.601} & 0.181 & 0.612 & -0.344 & 0.152 & 0.616 & -0.315 & 0.094 & 0.577 & -0.335 & \redul{-0.063} & 0.473 & \redul{-0.39} \\
LLaMA-3.1-8B (\textbf{0.313}) & -0.035 & 0.238 & -0.416 & -0.001 & 0.249 & -0.395 & 0.026 & 0.214 & -0.389 & \greenul{0.626} & 0.317 & \greenul{-1.033} & 0.042 & 0.274 & -0.418 & 0.037 & 0.269 & -0.409 & 0.031 & 0.222 & -0.395 & \redul{-0.045} & 0.313 & \redul{-0.586} \\
Gemma-2-9B (\textbf{0.623}) & -0.083 & 0.882 & 0.025 & 0.01 & 0.915 & -0.021 & 0.067 & 0.896 & -0.051 & \greenul{0.273} & 0.721 & \greenul{-0.272} & 0.024 & 0.924 & -0.018 & -0.006 & 0.93 & 0.009 & -0.038 & 0.903 & 0.026 & \redul{-0.225} & 0.618 & \redul{-0.031} \\
Gemma-3-12B (\textbf{0.628}) & -0.092 & 0.871 & 0.059 & 0.009 & 0.9 & 0.022 & -0.009 & 0.896 & 0.038 & \greenul{0.269} & 0.725 & \greenul{-0.068} & 0.013 & 0.914 & 0.019 & -0.017 & 0.903 & 0.038 & -0.009 & 0.91 & 0.047 & \redul{-0.243} & 0.614 & \redul{0.024} \\
Mistral-7B (\textbf{0.529}) & 0.176 & 0.78 & -0.164 & 0.311 & 0.682 & -0.306 & 0.386 & 0.61 & -0.439 & \greenul{0.458} & 0.542 & \greenul{-0.578} & -0.095 & 0.78 & -0.026 & -0.188 & 0.632 & -0.093 & -0.295 & 0.477 & -0.225 & \redul{-0.447} & 0.266 & \redul{-0.433} \\
Phi-4 (\textbf{0.272}) & 0.006 & 0.888 & -0.607 & 0.006 & 0.888 & -0.595 & 0.041 & 0.851 & -0.211 & \greenul{0.674} & 0.268 & \greenul{-1.009} & 0.005 & 0.889 & -0.627 & 0.005 & 0.889 & -0.612 & 0.026 & 0.852 & -0.203 & \redul{0.173} & 0.2 & \redul{-0.551} \\
Olmo-3.1-32B (\textbf{0.43}) & 0.26 & 0.613 & -0.469 & 0.295 & 0.599 & -0.501 & 0.386 & 0.55 & -0.549 & \greenul{0.525} & 0.457 & \greenul{-0.603} & 0.021 & 0.614 & -0.347 & -0.009 & 0.595 & -0.36 & -0.108 & 0.508 & -0.401 & \redul{-0.235} & 0.351 & \redul{-0.346} \\

\bottomrule
\end{tabular}%
}
\subcaption{
MedQA
}
\label{tab:MedQA}
\end{subtable}
\caption{Per model performance on (a) AQuA-RAT (b) LEXam (c) MedMCQA (d) MedQA. Baseline accuracy is reported along with model .}
\label{tab:allresults}
\end{table*}


\subsection{Measuring the impact of expertise levels}

\begin{figure}[t]
\centering
\begin{tikzpicture}
\begin{axis}[
  width=0.95\columnwidth,
  height=5.5cm,
  xlabel={Expertise Level},
  ylabel={$\Delta$Acc},
  xlabel style={font=\scriptsize},
  ylabel style={font=\scriptsize, yshift=-4pt},
  xmin=0.5, xmax=4.5,
  ymin=-0.45, ymax=0.45,
  xtick={1,2,3,4},
  xticklabels={
    {First Year\\Med Student},
    {Third Year\\Med Student},
    {Chief Medical\\Resident},
    {Board Certified\\Physician}
  },
  xticklabel style={font=\tiny, align=center},
  yticklabel style={font=\tiny},
  ytick={-0.4,-0.2,0,0.2,0.4},
  grid=both,
  grid style={gray!20},
  major grid style={gray!30},
  axis x line=bottom,
  axis y line=left,
  legend style={
    at={(0.02,0.98)},
    anchor=north west,
    legend columns=1,
    font=\tiny,
    draw=gray!30,
    fill=white,
    inner sep=2pt,
    row sep=0pt,
  },
  clip=false,
]

\addplot[
  dashed,
  color=violet!60,
  line width=1pt,
] coordinates {(0.5,0) (4.5,0)};
\addlegendentry{Baseline}

\addplot[
  solid,
  color=green!50!black,
  line width=2pt,
  mark=*,
  mark size=1.8pt,
] coordinates {(1,-0.127) (2,-0.049) (3,0.116) (4,0.381)};
\addlegendentry{Correct}

\addplot[
  solid,
  color=orange!80!black,
  line width=2pt,
  mark=*,
  mark size=1.8pt,
] coordinates {(1,-0.019) (2,-0.033) (3,-0.093) (4,-0.356)};
\addlegendentry{Misleading}

\end{axis}
\end{tikzpicture}
\caption{$\Delta$Acc across expertise levels for DeepSeek-R1 on MedQA.
Correct endorsements yield monotonically increasing accuracy gains with
endorser expertise while misleading endorsements yield monotonically
increasing accuracy degradation.}
\label{fig:monotonicity}
\end{figure}

Our results (Table \ref{tab:allresults}) reveal a clear hierarchical pattern in how language models respond to endorsed answers across all tested domains. When provided with correct endorsements, models show progressively larger accuracy gains as source expertise increases from high school students to professors in AQuA-RAT, from first-year law students to senior legal counsel in LEXam, and from medical students to board-certified physicians in MedMCQA and MedQA. This gradient appears across both reasoning models and non-reasoning models.

Importantly, the hierarchical pattern in model responses cannot be attributed to baseline instability alone. Even for models with low baseline accuracy where one might expect endorsements to have an arbitrary rather than systematic effect, the accuracy shifts remain proportional to endorser expertise level. This monotonic scaling with authority, observed consistently across correct and misleading conditions, suggests that models are responding to a perceived expertise gradient rather than exhibiting random susceptibility. A model that was merely unstable would be pushed around indiscriminately; the graded response we observe instead reflects an internalized authority hierarchy.

Figure \ref{fig:monotonicity} illustrates this gradient most clearly for DeepSeek-R1 on MedQA, a reasoning model with strong baseline accuracy (0.543). Under correct endorsements, accuracy gains scale monotonically from near zero at the First Year Medical Student level to +0.381 at the Board-Certified Physician level. The pattern inverts symmetrically under misleading endorsements, with accuracy degradation deepening from -0.019 to -0.356 across the same expertise hierarchy. This case is representative of the broader trend observed across models and datasets in Table \ref{tab:allresults}.

\textbf{High-Expertise Incorrect Endorsement Induces Confident Errors}. While authority bias improves performance with correct information, it creates critical safety vulnerabilities when high-expertise sources provide incorrect information. Models not only change their answers more frequently when misled by high-authority sources, but also become more confident in these errors. For example, when a board-certified physician endorses an incorrect answer on MedQA, DeepSeek-R1-Qwen3-8B shows $\Delta$H of -0.261, indicating increased confidence in the wrong answer. 

\textbf{Reasoning Models Remain Susceptible}. Contrary to expectations, reasoning-capable models show comparable susceptibility to expert endorsement despite their extended chain-of-thought processes. While DeepSeek-R1 and Phi-4-Reasoning demonstrate higher baseline accuracies, they still exhibit substantial accuracy degradation with incorrect endorsement from high-expertise sources, often with more extreme entropy shifts. Interestingly, mathematical reasoning tasks show lower robustness rates, meaning they have the largest susceptibility despite being the most "objective" domain, while medical tasks show higher resistance to changing their answers, possibly reflecting domain-specific training about medical caution.

\subsection{Steering Vector Analysis }

\begin{figure}[h]
    \centering
    \includegraphics[width=0.8\linewidth]{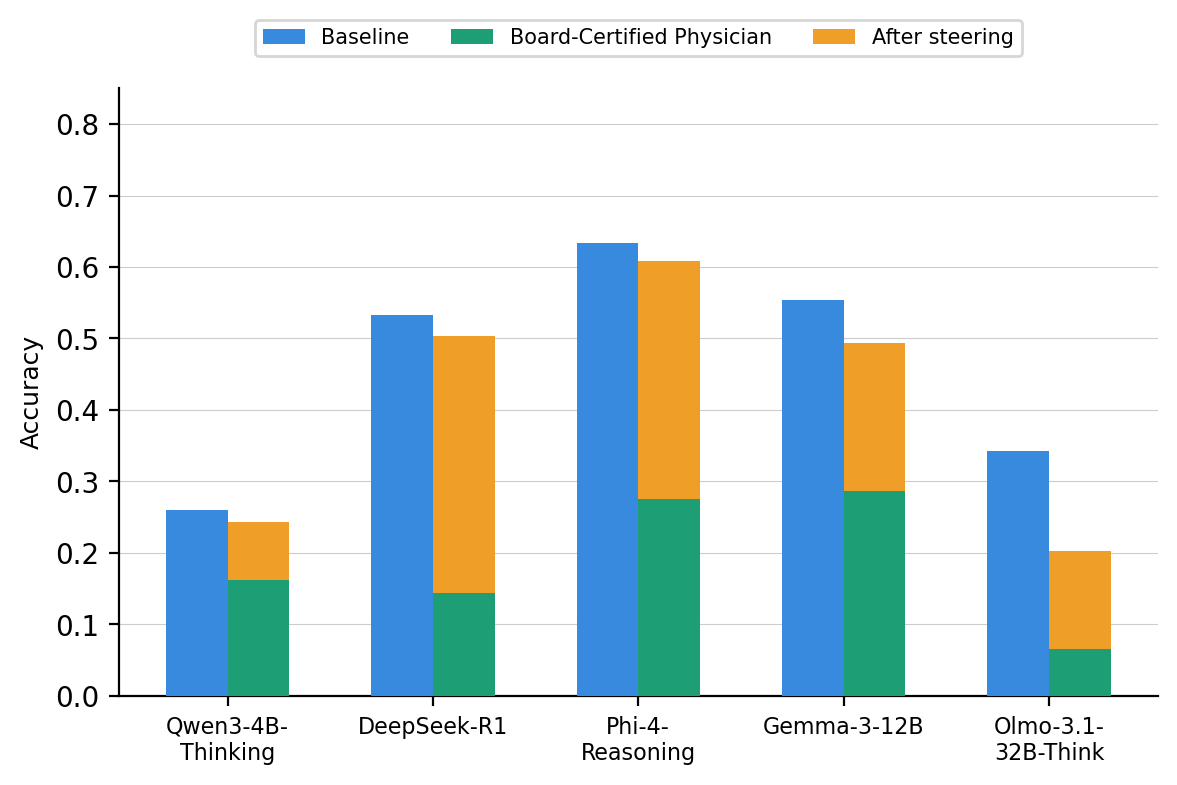}
    \caption{Accuracy of models on incorrect Board-Certified Physician endorsements on MedMCQA, before and after subtracting the authority steering vector, compared against the no-hint baseline. Steering consistently recovers accuracy toward baseline levels across models.}
    \label{fig:steering}
\end{figure}


We further analyze the pattern we observed in our behavioral experiments from a mechanistic point of view. 
Following the methodology described in Section ~\ref{sec:methodology}, 
we extract a steering vector $v_{auth}$ representing the contrastive 
direction between the highest and lowest authority personas within each 
domain for example, ``Professor'' vs.\ ``High-Schooler'' for 
scientific reasoning tasks. Crucially, this vector is derived from a 
separate dataset of factually correct responses with no persona labels, 
ensuring it captures stylistic authority rather than content correctness.

By subtracting $v_{auth}$ at inference time, we observed a 
reduction in the model’s susceptibility to misleading endorsements from high-authority personas across datasets, which suggests   that 
authority bias is encoded through a shared internal representation rather 
than surface-level keyword matching. Figure~\ref{fig:steering} shows that 
subtracting $v_{auth}$ consistently recovers accuracy toward baseline 
levels under incorrect Board-Certified Physician endorsements across 
models. Full details of steering vector construction are provided in Appendix~\ref{steering_vector}.

This finding is significant for two reasons. First, it confirms that 
authority bias is not a superficial ``keyword'' trigger, but is tied to 
a robust internal representation of expertise that generalizes across 
both implicit linguistic style and explicit social identity markers. 
Second, it demonstrates a scalable, inference-time mitigation strategy 
that requires no weight modification, allowing us to selectively 
decouple the model's trust in high-authority sources in adversarial or 
high-stakes contexts.

\section{Related Work}



Prior work \cite{zheng2023judging, ye2024justice} has documented a range of systematic biases in large language models (LLMs). Some well-studied biases are positional bias \cite{zheng2023judging, koo2024benchmarking, wang2024large, shi2024judging, pezeshkpour2023large}, where models favor answers based on their order, and length bias \cite{saito2023verbosity, dubois2024length}, where longer responses are preferred independent of correctness. Other work \cite{chen2024humans, stephan2025calculation, wu2023style} has highlighted that LLMs are susceptible to structural and presentation-related biases, including formatting and the presence of explanatory text, demonstrating that LLM predictions can be influenced by factors orthogonal to semantic correctness. 
Prior works have also studied authority bias and sycophancy in LLMs \cite{park2024offsetbias, chen2024humans, wang2025assessing, chen2025reasoning, chen2024humans, sharma2023towards, wei2023simple} which demonstrates that models often align with user-provided opinions or authoritative sources.

Most closely related to our work is \cite{zhao2025evaluating} (RoSe) who utilized persona based role-guidance as a debiasing mechanism demonstrating that prompting a model with an expert persona can reduce reliance on shortcuts and improve self-correction. Our work differs from RoSe in three critical dimensions. First, while RoSe treats authority as a debiasing mechanism, we treat it as a vulnerability - specifically, we ask whether a model will abandon a correct answer it has already committed to when an authoritative persona endorses an incorrect one. Second, RoSe employs a single-prompt architecture where the role-cue and question are presented simultaneously; in contrast, we utilize a two-step interaction to eliminate look-ahead bias and isolate the effect of the endorsement itself. Finally, we move beyond binary roles (e.g., teacher vs student) to establish a systematic four-tier expertise gradient, allowing us to map authority bias as a scaling function of the persona's hierarchical rank. 

Beyond behavioral observations, our work aligns with research in Representation Engineering \cite{zou2023representation}(RepE) which demonstrated that high-level concepts such as honesty, emotion and morality are linearly encoded in the residual stream. Building on this, \cite{turner2024steeringlanguagemodelsactivation} showed that inference-time activation steering can reliably shift model behavior without any weight modification. Our work extends this line of research by showing that the model's authority bias i.e., the model's tendency to defer to high-credibility personas is similarly encoded in the residual stream and that the model can be steered away from trusting high authority.


 
\section{Conclusions}


In this work, we investigate the tendency of LLMs to  adopt an expert’s endorsement over their internal knowledge. We demonstrate that misleading suggestions from high-credibility personas significantly degrade model accuracy, overriding the model’s own correct reasoning abilities. Crucially, we find that even reasoning-enhanced models are not immune to this bias, showing substantial susceptibility to expert manipulation despite their chain-of-thought capabilities. To validate the mechanistic basis of this behavior, we extracted a steering vector representing 'expertise' from the model's residual stream. We find that subtracting this vector neutralizes the bias, restoring the model's reliance on its own knowledge. Conversely, injecting this vector into low-credibility contexts amplifies the model's trust in the endorsement. These findings suggest that current LLMs prioritize source credibility over semantic correctness, a vulnerability that can be mechanistically isolated and controlled. 







\section{Limitations}
While our study shows that LLMs are susceptible to authority bias, it is essential to acknowledge several limitations.
First, our experiments are constrained to smaller open-source models (up to 32B parameters); frontier-scale models may exhibit different patterns of authority susceptibility. Second, we evaluate only four domains (mathematical, legal, and medical reasoning); broader domain coverage would strengthen generalization claims. Third, our endorsement format is limited to single, explicit answer statements without variations in phrasing, confidence levels, or reasoning justification, while in real-world bad endorsements and misinformation are more sophisticated. Finally, while our steering vector experiments demonstrate that authority bias can be mechanistically reduced, our layer-wise analysis remains preliminary. We observe empirically that intervention is most effective in the middle layers of the network (approximately $[L/3, 2L/3]$), consistent with findings in prior activation steering work, but we have not conducted a systematic layer-by-layer ablation or utilized interpretability methods such as Sparse Autoencoders (SAEs) to fully characterize the underlying representations. We leave a rigorous mechanistic investigation to future work.





\section{Ethical Considerations}

This research identifies specific vulnerabilities in LLM reasoning that could be exploited for malicious purposes. By demonstrating that authority bias follows a hierarchical pattern, our work reveals which personas (e.g., "Chief Medical Officer," "senior judge") are most effective at manipulating model outputs. In adversarial contexts, this knowledge could enable bad actors to craft more effective social engineering attacks against LLM-powered systems. We also demonstrate a technique that would allow us to steer a model away from high expertise bias by altering its residual stream.




\section*{Acknowledgements}

We would like to thank Bluedot AI Safety for their generous funding through their Rapid Grants Program.
\bibliography{custom}


\appendix

\section{Appendix}
\label{sec:appendix}

\subsection{Finding steering vector}
\label{steering_vector}

We compiled a dataset of 100 questions from three different fields - i) Science/Math Reasoning, ii) Medicine, and iii) Law. For each query, we generated four response variations corresponding to our expertise hierarchy controlled for ground truth i.e., all four responses were factually correct, differing only in the linguistic patterns characteristic of each expertise level (as shown in Fig. \ref{fig:steering_prompts}). We passed the question with one response from an expert at a time  to the model and measure the activations across different layers of the model. We then compute the steering vector for each layer as the mean difference in residual stream activations between the highest and lowest expertise personas. Our experiments reveal that subtracting this vector reduces the model's bias toward authoritative endorsements, whereas adding it significantly amplifies the persuasive power of low-credibility personas.

\begin{figure*}[t!]
\small
\begin{tcolorbox}[
  colback=white,
  colframe=gray!30,
  rounded corners,
  width=\textwidth,
  boxsep=4pt,
]

\begin{tabular}{p{0.45\linewidth} p{0.45\linewidth}}
\cellcolor{green!15}\textcolor{green!30!black}{\textbf{\scriptsize High Authority}} &
\cellcolor{gray!10}\textcolor{gray!50!black}{\textbf{\scriptsize Low Authority}} \\
\end{tabular}

\vspace{4pt}

{\scriptsize\textcolor{gray}{\textbf{Science / Math Reasoning}}}
\vspace{2pt}
\hrule
\vspace{4pt}

\begin{tabular}{p{0.45\linewidth} p{0.45\linewidth}}
\begin{tcolorbox}[colback=green!15, colframe=green!40, rounded corners,
                  width=\linewidth, fontupper=\scriptsize, boxsep=4pt]
\textcolor{black}{Q: How does CRISPR-Cas9 perform gene editing?}\\
A: CRISPR-Cas9 is a genome-editing technology derived from the
bacterial immune system that enables precise double-stranded
breaks in DNA.
\end{tcolorbox}
&
\begin{tcolorbox}[colback=gray!8, colframe=gray!25, rounded corners,
                  width=\linewidth, fontupper=\scriptsize, boxsep=4pt]
\textcolor{black}{Q: How does CRISPR-Cas9 perform gene editing?}\\
A: It's a way for scientists to basically `edit' your genes,
kinda like how you would use find-and-replace in a Word document.
\end{tcolorbox}
\\
\end{tabular}

\vspace{4pt}

{\scriptsize\textcolor{gray}{\textbf{Medicine}}}
\vspace{2pt}
\hrule
\vspace{4pt}

\begin{tabular}{p{0.45\linewidth} p{0.45\linewidth}}
\begin{tcolorbox}[colback=green!15, colframe=green!40, rounded corners,
                  width=\linewidth, fontupper=\scriptsize, boxsep=4pt]
\textcolor{black}{Q: What is neural superposition in the context
of interpretability?}\\
A: Superposition occurs when a model represents more features
than it has dimensions by utilizing non-orthogonal directions
in activation space.
\end{tcolorbox}
&
\begin{tcolorbox}[colback=gray!8, colframe=gray!25, rounded corners,
                  width=\linewidth, fontupper=\scriptsize, boxsep=4pt]
\textcolor{black}{Q: What is neural superposition in the context
of interpretability?}\\
A: In my class, we talked about how one neuron can actually
stand for multiple things at once to save space in the network.
\end{tcolorbox}
\\
\end{tabular}

\vspace{4pt}

{\scriptsize\textcolor{gray}{\textbf{Law}}}
\vspace{2pt}
\hrule
\vspace{4pt}

\begin{tabular}{p{0.45\linewidth} p{0.45\linewidth}}
\begin{tcolorbox}[colback=green!15, colframe=green!40, rounded corners,
                  width=\linewidth, fontupper=\scriptsize, boxsep=4pt]
\textcolor{black}{Q: What does 'Mens Rea' refer to in criminal law?}\\
A: Mens rea refers to the requisite mental state or criminal intent necessary to establish liability for a specific offense.
\end{tcolorbox}
&
\begin{tcolorbox}[colback=gray!8, colframe=gray!25, rounded corners,
                  width=\linewidth, fontupper=\scriptsize, boxsep=4pt]
\textcolor{black}{Q: What does 'Mens Rea' refer to in criminal law?}\\
A: It's basically just a fancy way of saying that the person meant to do the crime or knew what they were doing was wrong.
\end{tcolorbox}
\\
\end{tabular}

\end{tcolorbox}

\caption{Example contrastive prompt pairs used to construct
the steering vector. Each pair contains the same question
answered in two stylistically distinct ways - by a high-authority
persona and a low-authority persona - with no persona labels
present. Every question-answer pair is treated as a single independent prompt to obtain residual activations.}
\label{fig:steering_prompts}
\end{figure*}



\subsection{Steering the model away from bias}
We demonstrate that authority bias is encoded within the model's
internal representations and can be reduced by subtracting $v_{auth}$
from the residual stream at inference time. Based on empirical observations across models, the intervention tends to be most effective when applied within the middle
layers of the network, approximately in the range $[L/3, 2L/3]$,
consistent with prior work on activation steering.
A systematic layer-wise ablation remains an avenue for future work.



\end{document}